\PassOptionsToPackage{prologue,table}{xcolor}
\documentclass[sigconf,screen]{acmart}
\AtBeginDocument{%
  }

\setcopyright{acmlicensed}
\copyrightyear{2025}
\acmYear{2025}
\acmDOI{XXXXXXX.XXXXXXX}

\acmConference[Conference acronym 'XX]{Make sure to enter the correct
  conference title from your rights confirmation emai}{June 03--05,
  2025}{Dublin, Ireland}
\acmISBN{978-1-4503-XXXX-X/18/06}

\acmSubmissionID{37}

\usepackage{pifont}

\usepackage{float}

\usepackage{amsmath}
\usepackage{placeins}
\usepackage{amsfonts}
\usepackage{graphicx}
\usepackage{textcomp}
\usepackage{xcolor}
\usepackage{listings}
\usepackage{float} 
\usepackage{pythonhighlight}
\usepackage[T1]{fontenc}
\usepackage{verbatim}
\usepackage{url}
\usepackage{multirow}
\usepackage{booktabs} 
\usepackage{wrapfig}
\usepackage{tcolorbox}
\usepackage{algorithm}
\usepackage{algpseudocode}
\usepackage{amsmath}
 
\usepackage{float}
\usepackage{caption}
\usepackage{caption}
\usepackage{subcaption}
\usepackage{afterpage}
\usepackage{amssymb}  
 
\usepackage{xcolor}   
\usepackage{soul} 
\usepackage{xcolor} 
\sethlcolor{yellow} 
\renewcommand{\hl}[1]{#1}
\usepackage{booktabs}       
\usepackage{amsfonts}       
\usepackage{nicefrac}       
\usepackage{microtype}      
\usepackage{graphicx}       

\usepackage{enumitem}
\usepackage{colortbl}       
\usepackage[utf8]{inputenc}
\usepackage[T1]{fontenc}
\usepackage{graphicx}%
\usepackage{mathtools}
\usepackage{multirow}%
\usepackage{amsfonts}%
\usepackage{amsthm}%
\usepackage{mathrsfs}%
\usepackage[title]{appendix}%
\usepackage{xcolor}%
\usepackage{textcomp}%
\usepackage{manyfoot}%
\usepackage{booktabs}%
\usepackage{algorithm}%
\usepackage{algorithmicx}%
\usepackage{algpseudocode}%
\usepackage{listings}%

\usepackage{lineno}
\usepackage{xcolor}
\usepackage{pifont}
\newcommand{\cmark}{\ding{51}} 
\newcommand{\xmark}{\ding{55}} 
\definecolor{waymollgreen}{HTML}{CCFAEB} 
\definecolor{waymoblue}{HTML}{0077FF} 

\begin{document}




\title{Enhancing Low-Cost Video Editing with Lightweight Adaptors and Temporal-Aware Inversion}

\author{
  \begin{tabular}{c}
    Yangfan He$^{1*}$, Sida Li$^{2*}$, Jianhui Wang$^{3}$, Kun Li$^{4}$,\\ Xinyuan Song$^{5}$
    Xinhang Yuan$^{6}$, Kuan Lu$^{7}$, Menghao Huo$^{8}$,\\ Jingqun Tang$^{11}$, Yi Xin$^{12}$, Jiaqi Chen$^{9}$, Keqin Li$^{9}$,
    Miao Zhang$^{10}$, Xueqian Wang$^{10}$ \\[0.5em]
    $^{1}$ University of Minnesota—Twin Cities, 
    $^{2}$ Peking University,\\ 
    $^{3}$ University of Electronic Science and Technology of China, 
    $^{4}$ Xiamen University,\\
    $^{5}$ Emory University, 
    $^{6}$ Washington University, Saint Louis, 
    $^{7}$ Cornell University,
    $^{12}$ Nanjing University\\
    $^{8}$ Santa Clara University, 
    $^{9}$ Independent Researcher, 
    $^{10}$ Tsinghua University,
    $^{11}$ ByteDance Inc. 
  \end{tabular}
}

\renewcommand{\shortauthors}{Trovato et al.}
\settopmatter{printacmref=false} 

\begin{abstract}
Recent advancements in text-to-image (T2I) generation using diffusion models have enabled cost-effective video-editing applications by leveraging pre-trained models, eliminating the need for resource-intensive training. However, the frame independence of T2I generation often results in poor temporal consistency. Existing methods address this issue through temporal layer fine-tuning or inference-based temporal propagation, but these approaches suffer from high training costs or limited temporal coherence.
To address these challenges, we propose a  General and Efficient  Adapter (GE-Adapter) that integrates temporal, spatial and semantic consistency with Baliteral Denoising Diffusion Implicit Models (DDIM) inversion. This framework introduces three key components: (1) Frame-based Temporal Consistency Blocks (FTC Blocks) to capture frame-specific features and enforce smooth inter-frame transitions using temporally aware loss functions; (2) Channel-dependent Spatial Consistency Blocks (SCD Blocks) employing bilateral filters to enhance spatial coherence by reducing noise and artifacts; (3) a Token-based Semantic Consistency Module (TSC Module) to maintain semantic alignment through a combination of shared prompt tokens and frame-specific tokens. Extensive experiments on multiple datasets demonstrate that our method significantly improves perceptual quality, text-image relevance, and temporal coherence. The proposed approach offers a practical and efficient solution for text-to-video (T2V) editing. Our code is available in the supplementary materials.
\end{abstract}






\begin{teaserfigure}
  \centering
  \includegraphics[width=\textwidth]{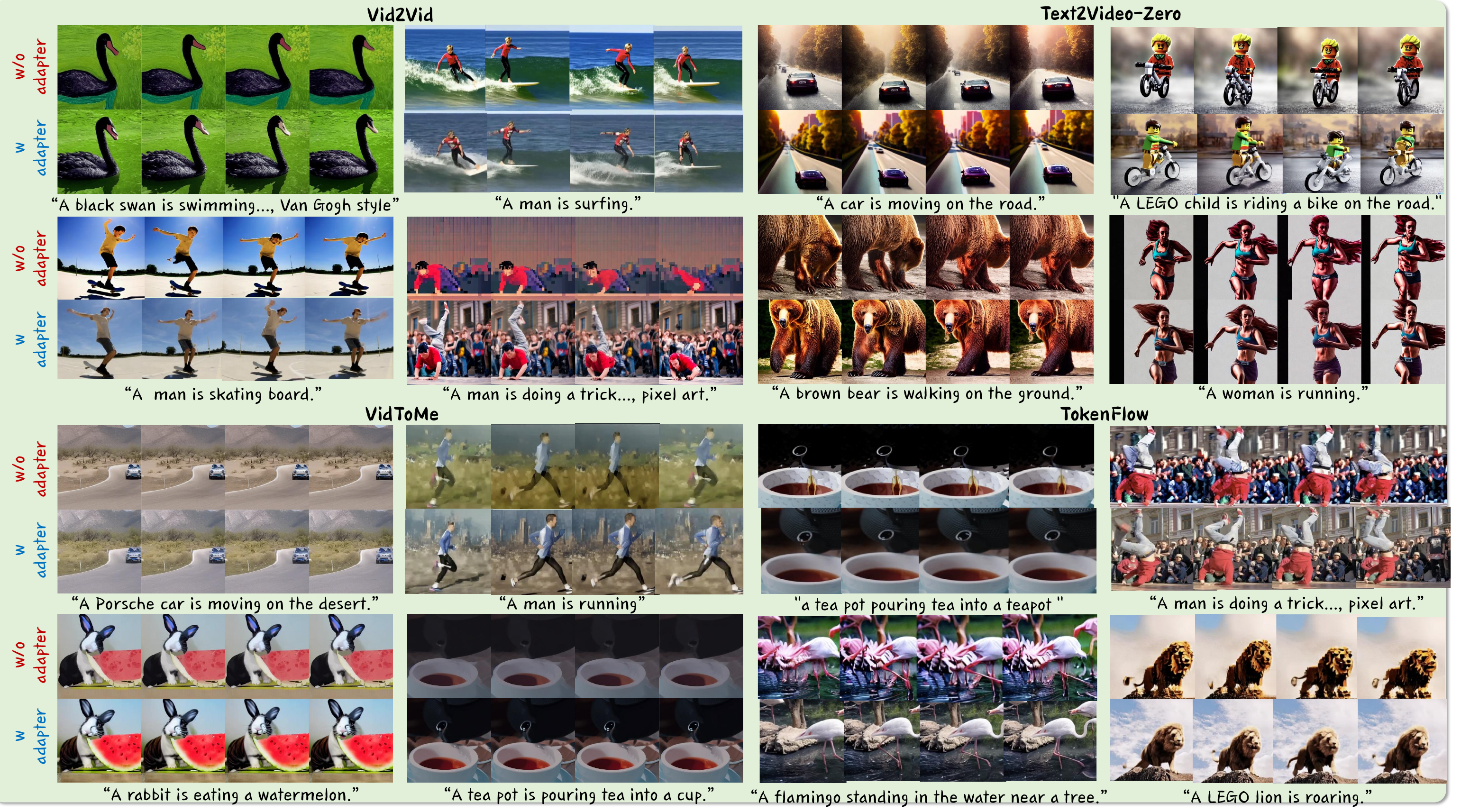}
  \caption{Visual comparison of generated video frames with and without our adapter applied to different algorithms.}
  \Description{Exa.}
  \label{fig:teaser}
\end{teaserfigure}

\maketitle


\def\thefootnote{*}\footnotetext{Equal contribution.}\def\thefootnote{\arabic{footnote}}

\section{Introduction}

\label{sec:intro}
Recent advances in deep learning have demonstrated remarkable progress across multiple domains, particularly in natural language processing, time series analysis, and computer vision applications~\cite{qiu2025easytime, qiu2025duet, qiu2024tfb, tao2024robustness, du2025zero, shen2024altgen,wang2025research, wang2025research, wang2025design, zhao2025optimizedpathplanninglogistics, xu2024autonomous, xu2024comet, weng2022large, zhong2025enhancing, li2025revolutionizing}. Building upon these foundations, text-to-video (T2V) generation and editing techniques~\cite{liu2024sorareviewbackgroundtechnology,wang2023modelscopetexttovideotechnicalreport,khachatryan2023text2videozerotexttoimagediffusionmodels,bartal2024lumierespacetimediffusionmodel,Tune-A-Video} have emerged as a cutting-edge research direction, focusing on synthesizing and manipulating dynamic video content that accurately reflects textual descriptions while maintaining visual coherence, spatial consistency, and temporal continuity. Recent advancements in diffusion models have significantly accelerated progress in T2V tasks, enabling more efficient and effective video generation and editing. Studies on parameter-efficient fine-tuning have demonstrated the potential of adapter-based methods \cite{xin2024v,xin2024vmt,xin2024parameter,yi2024towards,luo2024enhancing}

Traditional T2V methods rely on large-scale video datasets to learn spatial and temporal dynamics across diverse scenes. Pioneering works such as VideoGPT~\cite{yan2021videogpt}, CogVideo~\cite{hong2022cogvideo}, and Recipe~\cite{wang2024recipe} demonstrate the advantages of leveraging extensive datasets to capture fine-grained motion, object interactions, and scene transitions. However, these methods face significant challenges, including high computational costs, dataset biases, and the need for high-quality annotations.

To address these limitations, early approaches leveraging pre-trained text-to-image (T2I) diffusion models~\cite{wang2024ensembling, wang2024v, huo2024synthesizing, xu2024training} have emerged as a promising alternative due to their lower computational requirements. Notable examples include Animatediff~\cite{guo2023animatediff}, Tune-A-Video~\cite{wu2023tuneavideooneshottuningimage}, and TokenFlow~\cite{tokenflow2023}, which offer cost-effective solutions while maintaining acceptable performance. These approaches can be broadly categorized into training-based and training-free strategies.

Training-based strategies focus on fine-tuning temporal or attention layers to enhance temporal consistency and improve editing performance. While effective, this approach incurs high training costs and limited scalability. To mitigate these issues, adapter-based methods~\cite{mou2023t2iadapterlearningadaptersdig, balaji2023ediffitexttoimagediffusionmodels, chen2023visiontransformeradapterdense, feng2023trainingfreestructureddiffusionguidance} have been introduced. These methods integrate temporal or attention layers to reduce training overhead and improve generalization. However, their relatively large parameter sizes and limited adaptability highlight the need for further optimization. In contrast, training-free strategies, such as TokenFlow~\cite{geyer2023tokenflow}, propagate features between adjacent frames during inference. This eliminates training costs but can lead to lower-quality outputs.

To achieve a balance between computational efficiency and high-quality video generation, we propose a novel Consistency-Adapter Framework. This framework integrates temporal-spatial and semantic consistency while minimizing training expenses. A hierarchical temporal-spatial coherence module (HTC Module) enhances video quality through two key blocks: (1) Frame Similarity-based Temporal Consistency Blocks (FTC Blocks), which capture frame-specific information and reduce abrupt feature changes using temporally aware loss functions; and (2) Channel-Dependent Spatial Consistency Blocks (SCD Blocks), which use bilateral filtering to decrease noise and artifacts. Additionally, a Token-based Semantic Consistency Module (TSC Module) ensures semantic alignment by using shared prompt tokens for flexible editing and frame-specific tokens to maintain inter-frame consistency.

The main contributions of this paper are as follows:
\begin{itemize}[left = 0em]
    \item We propose a lightweight, plug-and-play consistency-adapter framework that balances computational efficiency and video quality by integrating temporal, spatial, and semantic consistency.
    \item We introduce a hierarchical temporal-spatial coherence module (HTC Module) featuring Frame Similarity-based Temporal Consistency Blocks (FTC Blocks) and Channel-Dependent Spatial Consistency Blocks (SCD Blocks), ensuring temporal and spatial coherence while reducing noise and artifacts. We also develop a token-based semantic consistency module (TSC Module) that employs both shared and frame-specific tokens to maintain semantic alignment across frames.
    \item We demonstrate that our lightweight adapter, with only 0.755M trainable parameters for the UNet portion and 15.4M for the prompt portion (total size: 860M), achieves over 50\% efficiency improvement compared to leading T2V models while enhancing temporal consistency, semantic alignment, and video quality.
\end{itemize}
\begin{figure*}[!ht]
    \centering
    \includegraphics[width=\textwidth]{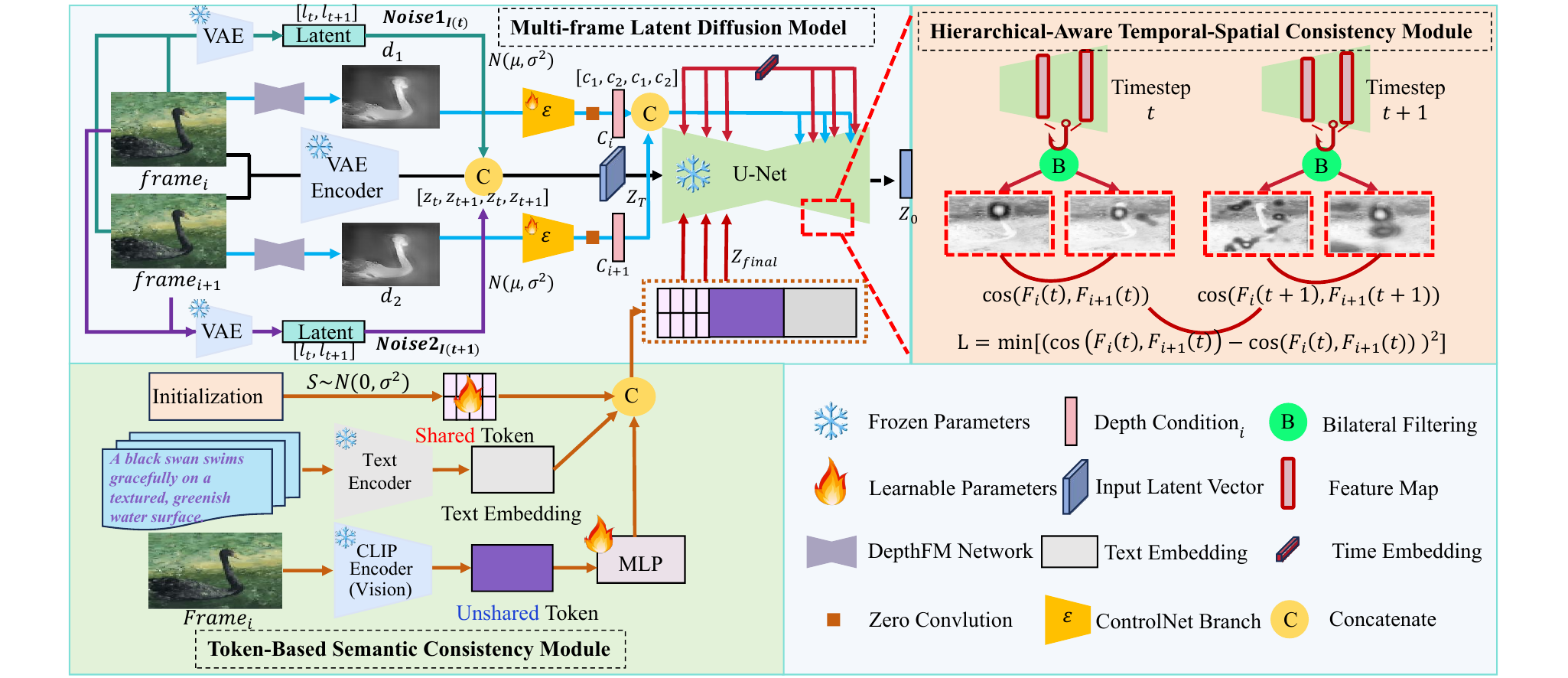} 
    \caption{The proposed framework integrates the Multi-frame Latent Diffusion Model, Hierarchical-Aware Temporal-Spatial Consistency Module, and Token-Based Semantic Consistency Module. The top-left illustrates Variational Autoencoder (VAE) encoding, Gaussian noise injection ($\mathrm{noise}_1$, $\mathrm{noise}_2$), and latent concatenation for UNet processing. The top-right shows the Temporal-Spatial Consistency Module, which optimizes frame transitions through a temporal loss. The bottom highlights the Semantic Consistency Module, which combines shared and unshared tokens to improve semantic alignment and reduce flickering, thereby enhancing both spatial and temporal coherence in video generation.}
    \label{fig:framework}
\Description{}\end{figure*}

\section{Related Work}
\label{sec:related_work}
\noindent\textbf{Text-to-Video Editing.}  
Text-to-Image (T2I) technology has achieved significant advancements through methods like Generative Adversarial Networks (GANs)~\cite{shen2023pbsl, NEURIPS2021_076ccd93, shen2023git, Karras_2020_CVPR, sauer2022styleganxlscalingstyleganlarge} and diffusion models~\cite{shen2024imagdressing, pan2018createtellgeneratingvideos, soviany2019imagedifficultycurriculumgenerative, shen2023advancing, shen2024boosting, gao2024exploring}. However, extending these advancements to Text-to-Video (T2V) remains challenging. Current T2V approaches include inversion and sampling methods, which optimize diffusion processes~\cite{rombach2022high, lu2022dpm, shen2024imagpose, salimans2022progressive}; upstream methods, which simplify fine-tuning with adapters~\cite{kim2024arbitrary}; and downstream methods, which enhance temporal and semantic consistency using complex attention mechanisms. 

For example, methods like Gen-L-Video~\cite{wang2023gen}, FLATTEN~\cite{cong2023flatten}, and StableVideo~\cite{chai2023stablevideo} focus on generating long videos with improved temporal coherence, while approaches like ControlVideo~\cite{zhao2023controlvideo} and MagicProp~\cite{yan2023magicprop} aim to enhance the quality of individual video frames. Despite these advancements, existing 2D U-Net-based T2V models often require training from scratch, freezing pre-trained T2I models and relying heavily on complex temporal layers, leading to computational inefficiency and temporal inconsistency. These methods lack a unified, efficient, and lightweight training paradigm that offers strong generalization with a single run and plug-and-play usability.

\noindent\textbf{DDIM Inversion for Enhanced Video Editing.}  
Denoising Diffusion Implicit Models (DDIM) Inversion~\cite{song2022denoisingdiffusionimplicitmodels} exploits the reversibility of DDIMs to control latent space content without regenerating the entire image. This technique enables precise adjustments to object shapes, styles, and details while maintaining consistency. Enhancements like EasyInv~\cite{zhang2024easyinvfastbetterddim} and ReNoise~\cite{garibi2024renoiserealimageinversion} refine inversion by iteratively adding and denoising noise, while Eta Inversion~\cite{kang2024etainversiondesigningoptimal} introduces a time- and region-dependent $\eta$ function to improve editing diversity. MasaCtrl~\cite{cao2023masactrltuningfreemutualselfattention} identifies object layouts by converting images into noise representations, and Portrait Diffusion~\cite{liu2023portraitdiffusiontrainingfreeface} merges Q, K, and V values for effective image blending. While these techniques excel in image editing, there remains a lack of optimized DDIM inversion methods tailored for video generation.

\noindent\textbf{Adapters for Video Editing.}  
Initially developed for natural language processing (NLP)~\cite{houlsby2019parameterefficienttransferlearningnlp}, adapters were introduced to efficiently fine-tune large pre-trained models, as demonstrated in BERT~\cite{devlin2019bertpretrainingdeepbidirectional} and GPT~\cite{radford2018improving}. In computer vision, adapters like ViT-Adapter~\cite{chen2023visiontransformeradapterdense} enable Vision Transformers (ViT) to handle diverse tasks with minimal fine-tuning. Similarly, ControlNet~\cite{zhang2023addingconditionalcontroltexttoimage} and T2I-Adapter~\cite{mou2023t2iadapterlearningadaptersdig} incorporate lightweight modules for diffusion models to provide additional control, while Uni-ControlNet~\cite{zhao2023unicontrolnetallinonecontroltexttoimage} reduces fine-tuning costs with multi-scale conditional adapters for localized control. However, current adapters remain limited to specific architectures and lack the flexibility and generalization required for robust video generation tasks.

\section{Method}
Text-to-video (T2V) editing requires preserving frame-to-frame temporal coherence, consistent semantics, and spatial structures. Given an input video and text prompts, the goal is to produce an edited video that: (1) \textbf{Frame Temporal Alignment}: Video editing should ensure temporal coherence among frames by creating smooth transitions between frames and accurately representing motion dynamics as described by the text prompt; (2) \textbf{Video Spatial Alignment}: The edited video should ensure that spatial structures and visual content remain consistent across all frames, aligned with the spatial details described in the text prompt while preserving the integrity of the original video’s regions; and (3) \textbf{Text-to-video Semantic Alignment}: The algorithm must also ensure that the edited video conveys the intended semantics described in the text prompt while maintaining the contextual consistency of the original video.

We propose a lightweight video adapter that promotes temporal and spatial consistency, as well as semantic alignment, while reducing training costs in 2D UNet-based T2V generation. Our approach integrates:
\begin{itemize}[left = 0em]
    \item A multi-frame latent diffusion model (Section~\ref{backbone});
    \item Frame Similarity-based Temporal Consistency Blocks (FTC Blocks, Section~\ref{sec:1}) to enforce temporal coherence across adjacent frames;
    \item Channel-dependent Spatial Consistency Blocks (SCD Blocks, Section~\ref{sec:2}) to stabilize noisy latents and reduce frame-level artifacts using bilateral filters;
    \item A Token-based Semantic Consistency Module (TSC Module, Section~\ref{sec:3}) to align text and video latents effectively.
\end{itemize}


\subsection{Base Diffusion model} \label{backbone}
This model differs from traditional methods by directly modeling latent features for consecutive video frames~\cite{khachatryan2023text2videozerotexttoimagediffusionmodels,qi2024deadiffefficientstylizationdiffusion, tokenflow2023}. In-depth analyses on the inner workings of diffusion models have also been reported~\cite{yi2024towards}, which support our design choices. By incorporating ControlNet branches~\cite{zhang2023addingconditionalcontroltexttoimage}, time embeddings~\cite{fraikin2024treprepresentationlearningtime}, and concatenated latent representations, the model aligns temporal features effectively, reducing flickering and structural inconsistencies. Additionally, the UNet’s decoder layers are optimized for high-resolution outputs, enabling fine-grained reconstruction of temporal dynamics and appearance details. Overall, the Multi-frame Latent Diffusion Model s serves as the backbone of our video editing framework, generating and aligning video frames through a diffusion process tailored for multi-frame consistency.

\begin{figure*}[!ht]
    \centering
    \includegraphics[width=\textwidth]{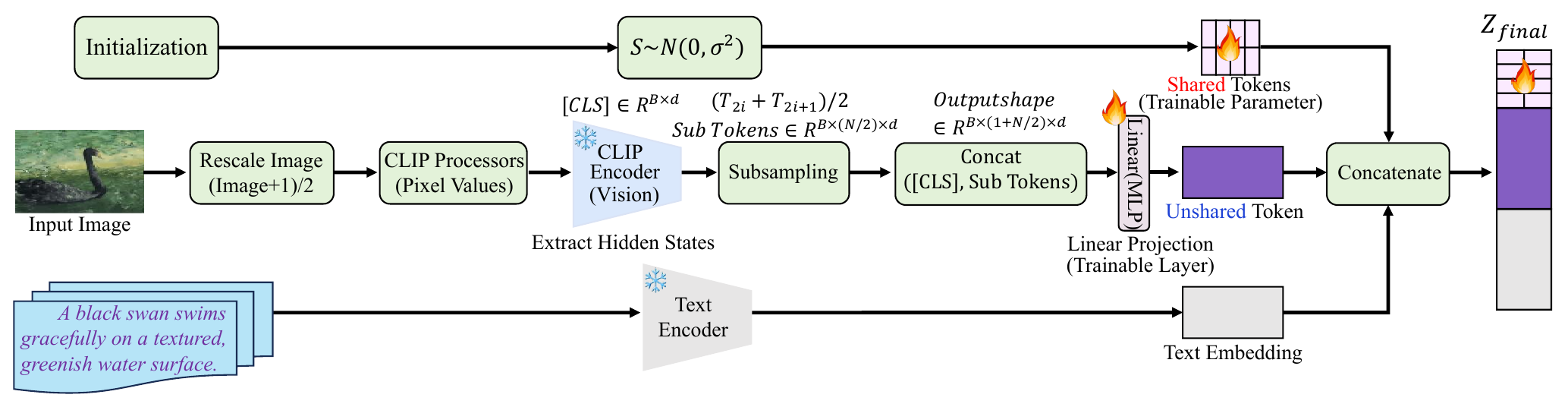} 
    \caption{This method combines shared and unshared tokens, maintaining a consistent shared token embedding across timesteps while updating frame-specific unshared tokens. This balances global context and local details. Guided by text embeddings, the diffusion model refines noise into semantically consistent images. Frame-dependent unshared tokens introduce per-frame variations in cross-attention guidance and input noise.}
    \label{fig:model_performance}

\Description{}\end{figure*}

The module begins by encoding consecutive video frames ($\mathrm{frame}_i$ and $\mathrm{frame}_{i+1}$) into latent representations using a VAE, as shown in Figure~\ref{fig:framework}. These latent encodings ($z_t, z_{t+1}$) are perturbed with Gaussian noise at each diffusion timestep to simulate noisy latent states ($\mathrm{noise}_1$ and $\mathrm{noise}_2$): $z_t \sim \mathcal{N}(\mu, \sigma^2)$. $\mu$ and $\sigma^2$ is the mean and variance of the Gaussian noise. 

In this model, $\mathrm{noise}_1$ and $\mathrm{noise}_2$ are applied separately to $z_t$ and $z_{t+1}$ to capture frame-specific variations such as lighting, texture, and motion, ensuring that each frame's unique features are preserved. By injecting independent noise distributions, the model prevents over-smoothing, retaining high-quality spatial details while maintaining temporal consistency. It also simulates realistic frame-to-frame degradations, which enhances the model’s ability to reconstruct natural and consistent videos during denoising.

The latent representations of consecutive frames, $z_t$ and $z_{t+1}$, along with their noise-injected counterparts, are then combined into a unified latent representation ($[z_t, z_{t+1}, z_t, z_{t+1}]$) to support cross-frame temporal modeling. This concatenated representation captures relationships between frames, enabling the model to jointly process both spatial and temporal features in the latent space. The noisy latent representations are further processed through a UNet-based architecture. Time embedding vectors, encoding the timestep information $t$, are introduced into the UNet to maintain structural consistency during the diffusion process. At each timestep, the latent representations are concatenated with control signals ($c_t, c_{t+1}, c_t, c_{t+1}$) generated by auxiliary ControlNet branches.



Latent representations for multiple frames are jointly processed through a combination of intermediate feature concatenation, time embedding integration, and latent concatenation. Temporal feature maps ($F_t, F_{t+1}$) are concatenated and injected into the UNet, enabling the model to capture inter-frame relationships effectively. Additionally, the timestep embeddings encoded within the UNet ensure that the noise schedule follows the temporal progression of the video. Furthermore, the concatenated latent representation ($[z_t, z_{t+1}, z_t, z_{t+1}]$) enhances the modeling of spatial-temporal dependencies between frames.

\subsection{Hierarchical-Aware Temporal-Spatial Consistency Module}

\begin{figure*}[!ht]
    \centering
    \includegraphics[width=\textwidth]{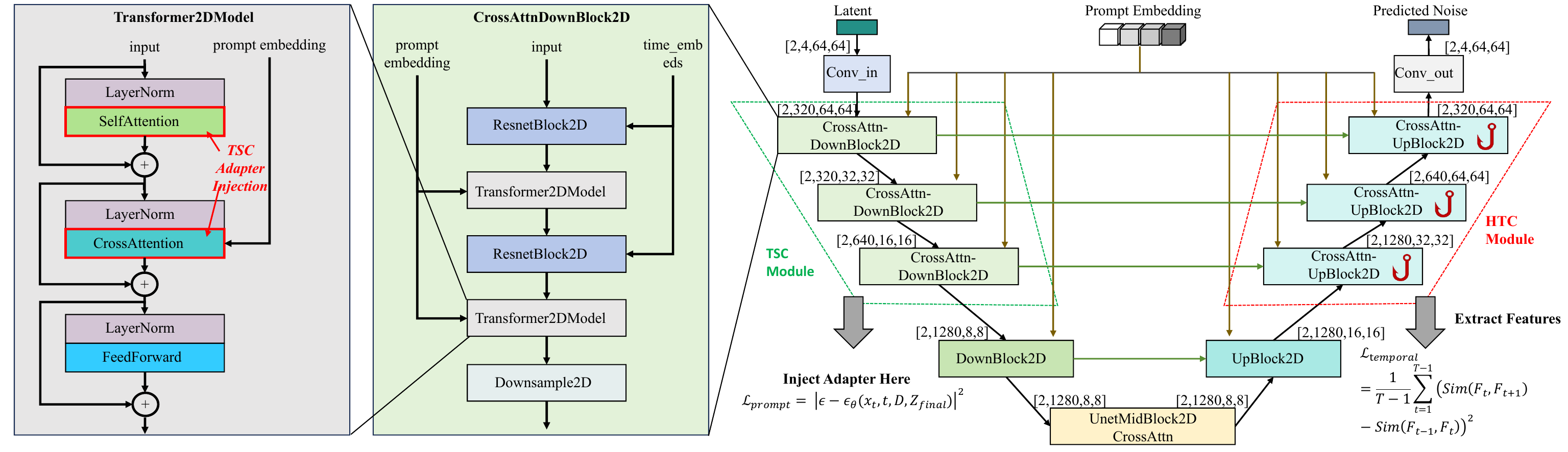} 
    \caption{Overview of the UNet framework, showing where TSC adapters and HTC modules are introduced for temporal coherence and feature extraction. The CrossAttnDownBlock2D and UpBlock2D use prompt embeddings and time embeddings, with adapters added for better prompt conditioning. The losses $L_{\text{prompt}}$ and $L_{\text{temporal}}$ guide robust noise prediction.}
    \label{fig:unet}
\Description{}\end{figure*}

\textbf{Frame Similarity-based Temporal Consistency Blocks} \label{sec:1}
Diffusion-based video generation methods have advanced significantly in recent years, but maintaining temporal consistency across video frames remains a critical challenge. Existing models attempt to incorporate temporal dimensionality into the diffusion process using techniques such as pseudo-3D convolutions \cite{singer2022makeavideotexttovideogenerationtextvideo}, sparse-causal attention \cite{wang2023zero}, and self-attention feature injection \cite{ceylan2023pix2video}. However, these approaches often face drawbacks such as high computational costs, limited scalability, and suboptimal generalization to diverse video content.
Unlike previous studies that primarily focus on large-scale retraining or post-hoc temporal smoothing, our approach introduces lightweight blocks with a novel temporal-aware loss function in the UNet's decoder layers. Details about these blocks are provided in Figure~\ref{fig:unet}:


Further details on these blocks are shown in Figure~\ref{fig:unet}:
\begin{equation}
x_{t+1} = x_t + \epsilon_t - \theta(x_t, t),
\end{equation}
where $x_t$ is the image at timestep $t$, $\epsilon_t$ is the predicted noise, and $\theta$ is the UNet model. This standard model produces static images and does not explicitly handle temporal relationships. We address this by incorporating lightweight, trainable adapters into the UNet, using hooks to extract intermediate feature maps $\mathbf{F}_{l,b}^{t}$ from each block $(l, b)$ at timestep $t$:
\begin{equation}
\mathbf{F}_{l,b}^{t} = \mathbf{W}_0 \mathbf{x} + \mathbf{B}_{l,b} \mathbf{A}_{l,b} \mathbf{x},
\end{equation}
where $\mathbf{x}$ is the input feature, and $\mathbf{B}_{l,b}$ and $\mathbf{A}_{l,b}$ are lightweight, learnable low-rank parameters.

To achieve smooth transitions between video frames, we use a similarity function to measure alignment between adjacent feature maps:
\begin{equation}
\text{Sim}(\mathbf{F}_t, \mathbf{F}_{t+1}) = \frac{\mathbf{F}_t \cdot \mathbf{F}_{t+1}}{\|\mathbf{F}_t\| \|\mathbf{F}_{t+1}\|}.
\end{equation}
We then define a temporal consistency loss:
\begin{equation}
\mathcal{L}_{\text{temporal}} = \frac{1}{T-1} \sum_{t=1}^{T-1} \Big( \text{Sim}(\mathbf{F}_t, \mathbf{F}_{t+1}) - \text{Sim}(\mathbf{F}_{t-1}, \mathbf{F}_t) \Big)^2,
\end{equation}
where $T$ is the total number of timesteps. By minimizing $\mathcal{L}_{\text{temporal}}$, the model aligns features between consecutive frames, reducing flickering and ensuring smoother transitions.

A standard diffusion loss is also required:
\begin{equation}
\mathcal{L}_{\text{diffusion}} = \mathbb{E}_{x_0, \epsilon, t} \Big[ \| \epsilon - \epsilon_\theta(x_t, t) \|^2 \Big],
\end{equation}
where $\epsilon$ is the noise added to $x_0$ at timestep $t$, and $\epsilon_\theta$ is the model’s predicted noise. The overall objective function combines the temporal consistency and diffusion losses:
\begin{equation}
\mathcal{L}_{\text{total}} = \lambda_{\text{temporal}} \mathcal{L}_{\text{temporal}} + \lambda_{\text{diffusion}} \mathcal{L}_{\text{diffusion}},
\end{equation}
where $\lambda_{\text{temporal}}$ and $\lambda_{\text{diffusion}}$ are set to 1 and 0.01, respectively. This balance promotes smooth, temporally consistent video editing while preserving denoising quality. 


\textbf{Channel-Dependent Spatially Consistent Denoising Blocks}\label{sec:2}
Another critical challenge in video editing is the inversion process, which is often necessary for generating edited outputs. Traditional  frame-by-frame DDIM inversion~\cite{qian2024siminversionsimpleframeworkinversionbased, wu2023tuneavideooneshottuningimage, ren2024customizeavideooneshotmotioncustomization} typically lack video-specific optimization, leading to inconsistencies across frames. To address this, we propose a bilateral filtering DDIM inversion technique that stabilizes latent representations and smooths spatial noisy latents without additional training, significantly improving frame-to-frame spatial consistency and ensuring seamless video generation. 

In Denoising Diffusion Implicit Models (DDIM) inversion for video generation, preserving consistency and quality across consecutive frames is difficult because of the inherent noise variations in the diffusion process. The reverse diffusion process denoises an input $ x_t $ according to:
\begin{equation}
p_\theta(x_{t-1} | x_t) = \mathcal{N}\bigl(x_{t-1}; \mu_\theta(x_t, t), \Sigma_\theta(t)\bigr),
\end{equation}
where $ \mu_\theta(x_t, t) $ is the predicted mean, and $ \Sigma_\theta(t) $ is the variance schedule that manages uncertainty at each reverse step.

Existing DDIM-based video inversion techniques face significant challenges in achieving frame-level smoothness and quality. The stochastic nature of the diffusion process often introduces uneven textures, noise artifacts, and loss of details, leading to visually inconsistent frames. These issues arise from the probabilistic framework of the reverse diffusion process, where the denoising of a noisy input $ x_t $ is governed by:  
\begin{equation}
x_{t-1} = \frac{1}{\sqrt{\alpha_t}} \left( x_t - \frac{1 - \alpha_t}{\sqrt{1 - \bar{\alpha}_t}} \epsilon_\theta(x_t, t) \right) + \sqrt{1 - \alpha_{t-1}} z,
\end{equation}
While $ \epsilon_\theta(x_t, t) $ predicts the noise, and $\alpha_t$, $\bar{\alpha}_t$ are scaling factors with $ z $ as sampled noise. We improve this framework with a bilateral filtering step applied to the noisy latents $ x_t $. It reduces artifacts by preserving edges and smoothing noise based on spatial and intensity features:
\begin{equation}
O_x = \frac{\sum_{y \in \mathcal{N}(x)} G_{\text{spatial}}(x, y) G_{\text{intensity}}(I_x, I_y) I_y}{\sum_{y \in \mathcal{N}(x)} G_{\text{spatial}}(x, y) G_{\text{intensity}}(I_x, I_y)},
\label{eq : video_inv}
\end{equation}
where $ \mathcal{N}(x) $ denotes the neighborhood of pixel $ x $, with $ y $ as neighboring pixels contributing to smoothing based on spatial proximity and intensity similarity, defined by their respective intensities $ I_x $ and $ I_y $. The spatial and intensity weights are calculated by:
\begin{equation}\label{eq:10}
G_{\text{spatial}}(x, y) = \exp\left(\frac{-(x - y)^2}{2\sigma_{\text{spatial}}^2}\right),
\end{equation}
\begin{equation}\label{eq:11}
G_{\text{intensity}}(I_x, I_y) = \exp\left(\frac{-(I_x - I_y)^2}{2\sigma_{\text{intensity}}^2}\right),
\end{equation}
where $ \sigma_{\text{spatial}} $ determines sensitivity to spatial distances, and $ \sigma_{\text{intensity}} $ controls the filter's response to intensity differences.

By incorporating bilateral filtering into the DDIM video inversion framework, the noisy latents $ x_t $ are smoothed at each timestep, producing refined latents $ x_t' $. The updated inversion step is:
\begin{equation}\label{eq:DDIM}
x_{t-1} = \frac{1}{\sqrt{\alpha_t}} \left( x_t' - \frac{1 - \alpha_t}{\sqrt{1 - \bar{\alpha}_t}} \epsilon_\theta(x_t', t) \right) + \sqrt{1 - \alpha_{t-1}} z,
\end{equation}
where $ x_t' $ is the filtered latent obtained from $ x_t $, ensuring smoother and more consistent intensity distributions. By replacing the original noisy latent with \( x_t' \), the denoising aligns with the smoothed latent distribution. This step significantly reduces noise artifacts and improves overall frame quality throughout the video inversion process.

\subsection{Token-Based Semantic Consistency Module}\label{sec:3}
Existing video editing algorithms often encounter difficulties in aligning video frames with text semantics because of static embeddings and fragmented text integration~\cite{singer2022makeavideotexttovideogenerationtextvideo, wang2023zero, ceylan2023pix2video}, frequently producing semantic inconsistencies and visual artifacts such as flickering. To address these issues, we propose a Token-Based Semantic Consistency Module that merges shared tokens for global context alignment with dynamic unshared tokens for frame-specific details. The shared tokens ensure that the main semantics of the text prompt remain consistent across all frames, maintaining a coherent narrative throughout the video. Meanwhile, the dynamic unshared tokens adapt to frame-specific variations, capturing localized details such as texture, lighting, or motion, which are crucial for preserving temporal continuity. This combination allows our module to balance global coherence with per-frame adaptability, effectively reducing artifacts like flickering and enhancing the overall quality of video editing outputs. The shared token embeddings $ T_{\text{share}} \in \mathbb{R}^{N_{\text{share}} \times 768} $ are initialized from a normal distribution $ \mathcal{N}(0, 0.02) $, more details see Figure~\ref{fig:model_performance}. For a given input image $ I $, the CLIP model's vision encoder extracts visual hidden features $ H_{\text{vision}} \in \mathbb{R}^{B \times N \times d} $, where $ B $ is the batch size, $ N = 50 $ is the sequence length, and $ d = 768 $ is the feature dimension. To construct a non-shared subset $ H_{\text{sub}} $, adjacent feature vectors along the second dimension are averaged, resulting in $ H_{\text{sub}} \in \mathbb{R}^{B \times N/2 \times d} $, defined as:
\begin{equation} 
\begin{aligned}
H_{\text{sub}}[:, i, :] &= \frac{H_{\text{vision}}[:, 2i, :] + H_{\text{vision}}[:, 2i+1, :]}{2}, \\
&\text{for } i \in \{0, 1, \dots, N/2 - 1\}.
\end{aligned}
\end{equation}

Next, a projection matrix $ W \in \mathbb{R}^{d \times d} $ is applied to $ H_{\text{sub}} $ to preserve the feature dimension:
\begin{equation}
Z_{\text{unshare}} = H_{\text{sub}} \cdot W, \quad W = \text{nn.Linear}(d, d).
\end{equation}
The final text embedding for temporal-aware fine-tuning is constructed as:
\begin{equation}
Z_{\text{final}} = \left[ T_{\text{share}}; Z_{\text{frame}}; \mathcal{C}(Z) \right],
\end{equation}
where $ T_{\text{share}} $ represents the shared token embedding, $ Z_{\text{frame}} $ is the frame-specific unshared token, and $ \mathcal{C}(Z) $ concatenates conditional and unconditional embeddings along the first sequence dimension. The concatenation is denoted by $[ \cdot ]$.

During the denoising process, cross-attention provides text guidance by mapping the latent features $ X_t \in \mathbb{R}^{M \times d} $ to updated features $\tilde{X}_t$ using the final text embedding $ Z_{\text{final}} \in \mathbb{R}^{L \times d} $ as keys and values:
\begin{equation}
Q = W_Q^\top X_t,\quad K = W_K^\top Z_{\text{final}},\quad V = W_V^\top Z_{\text{final}},
\end{equation}
\begin{equation}\label{eq:18}
\tilde{X}_t = \text{softmax}\left(\frac{QK^\top}{\sqrt{d}}\right)V.
\end{equation}
where $ W_Q \in \mathbb{R}^{M \times M} $, $ W_K \in \mathbb{R}^{L \times d} $, and $ W_V \in \mathbb{R}^{L \times d} $ are the learnable projection matrices for the query, key, and value transformations, respectively.

The updated cross-attention map $\tilde{X}_t$ is integrated into the noise prediction function $\epsilon_{\theta}$ to guide the denoising process. The denoising step at timestep $ t $ can then be expressed as:  
\begin{equation}
x_{t-1} = x_t - \alpha_t\epsilon_{\theta}(x_t, \tilde{X}_t),
\end{equation}
where $ x_t \in \mathbb{R}^{M \times d} $ (consistent with $ X_t $ and $\tilde{X}_t)$ is the noisy latent at step $ t $. The final text embedding $ Z_{\text{final}} $ integrates shared, frame-specific, and conditional/unconditional embeddings to compute $\tilde{X}_t$, aligning the denoising operation with frame-specific semantics to ensure global consistency and preserve local details.

During training, the parameters of the CLIP vision encoder ($\theta$) remain frozen, while the adapter parameters, shared embeddings, and projection layers for unshared tokens ($\phi$) are optimized iteratively to minimize the following loss function:
\begin{equation}
\resizebox{\linewidth}{!}{$
\text{Loss} = 
\begin{cases}
\left | \epsilon - \epsilon_{\theta}(x_{t}, t, D, Z_{\text{final}}) \right |^{2}, & \text{for } t \in [0, 0.5T], \\
\left | \epsilon - \epsilon_{\theta}(x_{t}, t, D, Z_{\text{final}}) \right |^{2} + \lambda \mathcal{L}_{\text{temproal}}, & \text{for } t \in [0.5T, T]
\end{cases}
$}
\end{equation}
where $\epsilon$ represents the noise at timestep $t$, $x_t$ is the input state, and $\mathcal{L}_{\text{temproal aware}}$ is the adjacent frames constraint. The adapter's parameters are updated as follows:
\begin{equation}\label{eq:gradient}
\Theta_{k+1} = \Theta_{k} - \eta \nabla_{\Theta} \text{Loss}(\Theta_{k})
\end{equation}
With $ \Theta = \{\phi_{\text{adapter}}, \phi_{\text{unshared}}, T_{\text{share}}\} $ representing the adapter, unshared token, and shared token embedding parameters, and $\eta$ as the learning rate, the adapter ($\phi_{\text{adapter}}$) remains active only during the extended training interval from 0.5 to 1.0, where it influences $\nabla_{\Theta} \text{Loss}$, otherwise remaining inactive.


\section{Experiments}
\subsection{Implementation Details}
We trained a Stable Diffusion v1.5-based model with an 860M-parameter UNet and a 123M-parameter text encoder. We integrated a ControlNet specialized for depth data, coupled with an adapter that included a jointly trained PrefixToken module to enhance video-prompt alignment. Training was conducted in mixed precision (fp16) with a learning rate of 3e-5 and an input frame resolution of 512×512. The 1.3G-parameter ControlNet was trained for 20 hours using four RTX4090 GPUs. The PrefixToken module comprises 2.3M parameters, and when jointly trained with the UNet adapter, it requires 18 hours using one RTX4090 GPU. The same hyperparameters were employed during ControlNet/PrefixToken training. We used the AdamW optimizer with \(\beta_1=0.9\) and \(\beta_2=0.999\). The LoRA update matrix had a dimension of 4. The learning rate was 3e-5, with 500 warm-up steps in the scheduler, and the gradient accumulation steps were 8.\\
\textbf{UNet adapter.} UNet adapter only tunes attentions layer in UNet downsampling block.The adapter parameter size is 860 M. Unet adapter and prompt adapter train together, needs 18 hours with only one RTX4090 GPU. \\
\textbf{Prompt adapter.} The prompt learner contains a trainable sharetoken and layers that map image information to tokens. The adapter parameter size is 123 M. \\
\textbf{Controlnet.} Controlnet initialized from Unet. The model parameter size is 1.3G. It trained on MSRVTT dataset for 20 hours with four RTX4090 GPUs.

\subsection{Dataset}
We used the MSR-VTT dataset~\cite{7780940}, which has 10,000 video clips across 20 categories and 20 English captions each, resulting in about 29,000 unique words. The dataset is split into 6,513 training, 497 validation, and 2,990 test clips. We modified captions using the OpenAI ChatGPT API to produce additional variations and employed a specialized DataLoader to batch adjacent frames in early denoising steps. The MP4 videos were converted to WebDataset format for faster training. During inference, adapter weights were loaded into the StableDiffusionPipeline and merged at a 50\% ratio.

\begin{table}[!ht]
\centering
\caption{Comparison of algorithms with and without temporal awareness was conducted using LPIPS, CLIP, and FID metrics on the MSR-VTT testing set. Adapter parameters were applied during the 0.9-1.0 time intervals of the denoising process. Since text-to-video generation relies solely on textual input without any original source video, there is no FID-based distance between real and generated frames.}
\label{tab:algorithm_comparison}
\resizebox{0.95\linewidth}{!}{%
\begin{tabular}{l c c c c}
\toprule
\textbf{Algorithm} & \textbf{Adapter} & \textbf{LPIPS $\downarrow$} & \textbf{CLIP $\uparrow$} & \textbf{FID $\downarrow$} \\
\midrule

\multirow{2}{*}{Text2Video-Zero \cite{khachatryan2023text2videozerotexttoimagediffusionmodels}} 
 & \cellcolor{waymollgreen}\cmark & \textcolor{waymoblue}{\textbf{0.319}} & \textcolor{waymoblue}{\textbf{31.38}} & -- \\
 & \xmark & 0.402 & 28.82 & -- \\
\midrule

\multirow{2}{*}{TokenFlow \cite{geyer2023tokenflowconsistentdiffusionfeatures}} 
 & \cellcolor{waymollgreen}\cmark & \textcolor{waymoblue}{\textbf{0.129}} & \textcolor{waymoblue}{\textbf{29.15}} & \textcolor{waymoblue}{\textbf{143.21}} \\
 & \xmark & 0.135 & 28.90 & 166.72 \\
\midrule

\multirow{2}{*}{Vid2Vid \cite{wang2018videotovideosynthesis}} 
 & \cellcolor{waymollgreen}\cmark & \textcolor{waymoblue}{\textbf{0.253}} & \textcolor{waymoblue}{\textbf{32.78}} & 154.30 \\
 & \xmark & 0.303 & 31.94 & \textcolor{waymoblue}{\textbf{149.40}} \\
\midrule

\multirow{2}{*}{VidToMe \cite{li2023vidtomevideotokenmerging}} 
 & \cellcolor{waymollgreen}\cmark & \textcolor{waymoblue}{\textbf{0.122}} & \textcolor{waymoblue}{\textbf{29.94}} & \textcolor{waymoblue}{\textbf{143.33}} \\
 & \xmark & 0.123 & 29.93 & 146.27 \\
\bottomrule
\end{tabular}%
}

\end{table}

\subsection{Main Results}
\textbf{Superior Performance Across Algorithms and Settings} 
Table~\ref{tab:ablation_controlnet_individual_components} shows that our adapter is effective with different base diffusion models (e.g., SD-XL) and ControlNet conditioning modes. For instance, with Canny Edge conditioning, FID improves from 343.21 to 338.78, and LPIPS from 0.729 to 0.725. With Human Pose conditioning, CLIP scores increase from 29.49 to 33.56, while FID improves from 365.91 to 362.83 and LPIPS from 0.763 to 0.746, indicating enhanced semantic alignment. With Depth Map conditioning, FID decreases from 343.33 to 339.10, LPIPS from 0.721 to 0.718, and CLIP rises from 29.89 to 31.67. These results demonstrate the adapter’s ability to improve visual quality, semantic alignment, and frame-to-frame coherence under diverse conditions.

Table~\ref{tab:algorithm_comparison} highlights the compatibility of our adapter with various T2I-based T2V algorithms, showcasing its ability to consistently enhance performance across diverse methods~\cite{tokenflow2023,chu2024medm,li2023vidtomevideotokenmerging,khachatryan2023text2videozerotexttoimagediffusionmodels} with and without adapter injection demonstrates notable improvements in perceptual quality and text-image alignment: Text2Video-Zero achieves smoother transitions and better prompt adherence with LPIPS reduced from 0.402 to 0.319 and CLIP increased from 28.82 to 31.38; TokenFlow improves motion and texture consistency with a FID drop from 166.72 to 143.21, alongside LPIPS and CLIP gains; Vid2Vid enhances motion alignment with LPIPS decreasing from 0.303 to 0.253 and CLIP increasing from 31.94 to 32.78; and VidToMe improves perceptual quality and semantic alignment with LPIPS reduced from 0.126 to 0.114 and CLIP rising from 25.00 to 28.12, indicating enhanced perceptual quality and semantic alignment. 

\section{Ablation Study}
\textbf{Ablation of Token-Based Semantic Consistency Module.}
Our ablation study confirms that shared tokens are essential for maintaining global semantic alignment. When the adapter model (including shared tokens) is used, it outperforms the Base model by keeping frame-level semantic focus on key elements such as a “swan” or “water,” thus reducing attention drift. This also enhances details, for instance in neck movements or water ripples, and provides smooth transitions that tie motion and ripple effects together for stronger semantic consistency. Removing shared tokens severely degrades semantic alignment, as shown in Figure~\ref{fig:token_comparison} (in supplementary material), where TokenFlow's CLIP score drops from 29.4 to 20.1, FID increases from 135.7 to 387.8, and LPIPS rises from 0.14 to 0.17, disrupting global coherence and local detail quality, as further visualized in Figure~\ref{fig:token_ablation} (in supplementary material), with the lion's shape and textures becoming unrecognizable.


Boosting unshared tokens helps preserve attributes of the original video but can limit the model’s flexibility for novel textual prompts. For example, TokenFlow struggles with “A lion is walking on the grass,” and Text2Video with “A man is running,” when unshared tokens are too heavily weighted.


Figures~\ref{fig:token_comparison} and~\ref{fig:token_ablation} 12 (all in supplementary material) underline that too many unshared tokens degrade both global semantics and frame transitions. TokenFlow’s CLIP score, for instance, declines to 27.8, while FID and LPIPS deteriorate. A suitable mix of shared and unshared tokens is thus necessary for both fine detail and broad semantic integrity. We also vary the number of shared tokens used in the Prompt Adapter; visual results in Figure~\ref{fig:token_appendix} (in supplementary material) and metric outcomes indicate that 18 shared tokens offer an effective balance. Excessive shared tokens risk losing text-specific information.


\definecolor{myred}{RGB}{255, 0, 0}    
\definecolor{mygreen}{RGB}{0, 200, 0}  
\definecolor{myyellow}{RGB}{153, 102, 0} 
The activation time period of the Unet adapter is crucial, as shown in Figure~\ref{fig:ts} (in supplementary material), which evaluates the effects of temporal-aware loss training ranges and adapter activation ranges during inference. While a 0.5-1.0 training range combined with a 0.5-1.0 inference range achieves the best metrics across CLIP, FID, and LPIPS, Figure~\ref{fig:stepvis} (in supplementary material) reveals trade-offs: the broader 0.5-1.0 range introduces excessive constraints, resulting in over-smoothed details and visual blurriness. To resolve this, a narrower 0.9-1.0 inference range was adopted while keeping the broader 0.5-1.0 training range, striking a balance between temporal consistency and sharp, clear visual outputs.
\begin{table}[htbp]
\centering
\LARGE 
\small
\caption{An ablation study on pretrained ControlNet conditions (controlnet-canny, openpose, depth-sdxl-1.0) with adapter results (\hl{yellow}) on MSR-VTT human-edited cases for Human Pose demonstrates compatibility and performance gains, with enabled components in \cellcolor{green}{green}.}
\renewcommand{\arraystretch}{1.3}  
\resizebox{\linewidth}{!}{%
\begin{tabular}{c|c|c|c|c|c}
\toprule
\textbf{Canny Edge} & \textbf{Human Pose} & \textbf{Depth Map} & \textbf{FID $\downarrow$} & \textbf{CLIP $\uparrow$} & \textbf{LPIPS $\downarrow$} \\
\midrule
\cellcolor{green}\cmark & \xmark & \xmark & 343.21 (\hl{338.78}) & 29.15 (\hl{31.44}) & 0.729 (\hl{0.725}) \\
\xmark & \cellcolor{green}\cmark & \xmark & 365.91 (\hl{362.83}) & 29.49 (\hl{33.56}) & 0.763 (\hl{0.746}) \\
\xmark & \xmark & \cellcolor{green}\cmark & 343.33 (\hl{339.10}) & 29.89 (\hl{31.67}) & 0.721 (\hl{0.718}) \\
\bottomrule
\end{tabular}%
}
\label{tab:ablation_controlnet_individual_components}

\end{table}
\begin{table}[htbp]
\centering
\caption{Ablation Study: Effects of Inv on Different Algorithms.}
\resizebox{0.9\linewidth}{!}{%
\begin{tabular}{c|c|ccc}
\toprule
\textbf{Algorithm} & \textbf{Bilateral Inv} & \textbf{FID $\downarrow$} & \textbf{CLIP $\uparrow$} & \textbf{LPIPS $\downarrow$} \\
\midrule
\multirow{2}{*}{Tokenflow} & \xmark & 135.66 & 29.17 & 0.142 \\
                           & \cellcolor{waymollgreen}\checkmark & \textbf{\textcolor{waymoblue}{122.52}} & \textbf{\textcolor{waymoblue}{29.35}} & \textbf{\textcolor{waymoblue}{0.136}} \\
\midrule
\multirow{2}{*}{Vid2Vid} & \xmark & 387.76 & 20.08 & 0.210 \\
                          & \cellcolor{waymollgreen}\checkmark & \textbf{\textcolor{waymoblue}{320.80}} & \textbf{\textcolor{waymoblue}{25.84}} & \textbf{\textcolor{waymoblue}{0.166}} \\
\midrule
\multirow{2}{*}{VidToMe} & \xmark & 357.25 & 25.00 & 0.126 \\
                          & \cellcolor{waymollgreen}\checkmark & \textbf{\textcolor{waymoblue}{143.72}} & \textbf{\textcolor{waymoblue}{28.12}} & \textbf{\textcolor{waymoblue}{0.114}} \\
\midrule
\multirow{2}{*}{Text2Video} & \xmark & -- & 29.10 & 0.347 \\
                             & \cellcolor{waymollgreen}\checkmark & -- & \textbf{\textcolor{waymoblue}{30.07}} & \textbf{\textcolor{waymoblue}{0.244}} \\
\bottomrule
\end{tabular}%
}
\label{tab:ablation_inv}

\end{table}

\noindent\textbf{Ablation of Hierarchical-Aware Temporal-Spatial Consistency Module} 
We examine the effect of bilateral filtering within the Channel-Dependent Spatially Consistent Denoising Blocks. Figure~\ref{fig:bilateral_filtering} (in supplementary material) demonstrates that bilateral filtering reduces blur and jitter in generated frames, improving realism for objects such as penguins and rabbits. In Table~\ref{tab:ablation_inv}, TokenFlow’s FID improves from 135.66 to 122.52, CLIP rises from 29.17 to 29.35, and LPIPS drops from 0.142 to 0.136, showing tangible gains in perceptual quality and coherence.

\begin{figure}[!ht]
    \centering
    \includegraphics[width=\linewidth]{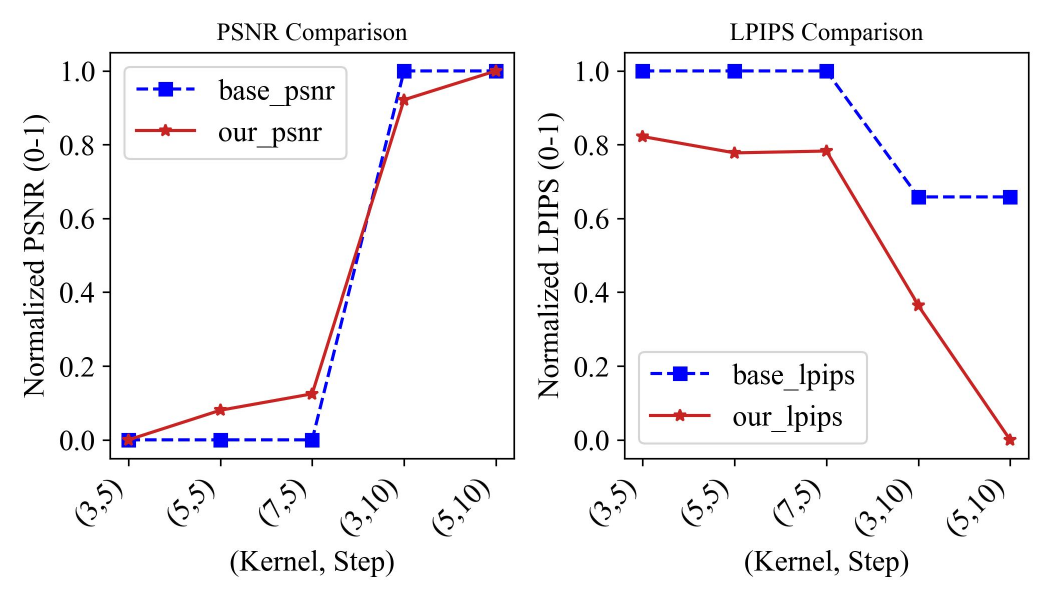}
    \caption{Comparison of step and kernel sizes $N(\cdot)$ within bilateral filtering inversion in the Stable Diffusion pipeline. The observed trends are consistent with those in vid2vid and three other algorithms (not shown due to space limitations).}
    \label{fig:psnr_ssim_comparison}
\Description{}\end{figure}

We also explored the advantages of our inversion method at smaller step sizes, as shown in Figure~\ref{fig:psnr_ssim_comparison}(in supplementary material), particularly in the 3-5 step configuration. At these smaller step sizes, our inversion method significantly improves temporal coherence, resulting in smoother transitions between frames. This is especially crucial in video generation, where smaller step sizes reduce noise and enhance fine details by minimizing artifacts like flickering or blurry transitions that often occur at larger step sizes. As a result, the model demonstrates more accurate frame-to-frame consistency, with enhanced details and smoother motion, underscoring the effectiveness of our inversion method at smaller step sizes. 

Furthermore, as is shown in Figure~\ref{fig:psnr_ssim_comparison}, we also observed that by appropriately increasing the kernel size (and its corresponding steps), image quality metrics (e.g., PSNR) can be enhanced while simultaneously improving coherence (e.g., LPIPS), ultimately producing clearer and more coherent visual outputs.

\subsection{User Study}
\begin{figure}[!ht]
    \centering
    \includegraphics[width=\linewidth]{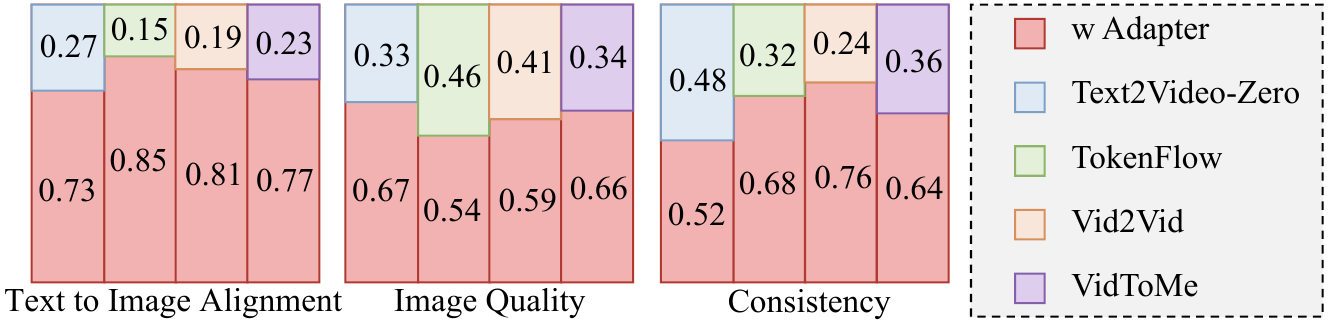}
    \caption{User study comparison of video generation algorithms with and without adding our adapter.}
    \label{fig:user_study}

\Description{}\end{figure}

We selected 67 random participants with diverse genders, ages, and educational backgrounds. They were asked to evaluate video outputs based on three dimensions: the coherence between frames, the alignment between text and frames, and the quality of the video frames themselves. For each algorithm, we selected 10 video editing cases with an adapter and 10 without, forming 20 pairs of videos in total. Each pair included one video generated with the adapter and one without. The participants were unaware of which videos had the adapter applied and were informed only that the videos were generated using different algorithms. They were instructed to choose the video they considered the best in each pair. Afterward, we collected their preferences for videos with and without adapters across all four algorithms, along with their selections in the three evaluation dimensions of text-to-image alignment, image quality, and consistency. Finally, we averaged the data from the 67 participants to calculate overall preference proportions for each algorithm and evaluation dimension. Figure~\ref{fig:user_study} shows adapter-enhanced videos preferred across algorithms, with notable gains in consistency for VidToMe and image quality for TokenFlow, enhancing overall alignment and frame quality.

More experiment details and the qualitative results are in the supplementary material.

\section{Conclusion}
We introduced a prompt-learning adapter, GE-Adapter, to improve temporal consistency and visual quality in text-guided video editing using pre-trained text-to-image diffusion models. Our plug-and-play adapter integrates temporal, spatial, and semantic consistency with Baliteral DDIM inversion, reducing flickering and refining text-to-video alignment at minimal training cost. Additionally, it incorporates three key components—Frame-based Temporal Consistency Blocks (FTC Blocks), Channel-dependent Spatial Consistency Blocks (SCD Blocks), and a Token-based Semantic Consistency Module (TSC Module)—to enhance perceptual quality and text-image relevance. This approach is also compatible with diverse video editing systems.

\bibliographystyle{ACM-Reference-Format} 
\bibliography{references} 

\clearpage
\appendix


\section{Qualitative Results}
Figure~\ref{fig:lora_comparison} (in supplementary material) highlights the \textbf{crucial} role of the adapter in enhancing video quality across a range of scenarios and algorithms. In the “lion roaring” example, videos without the adapter show inconsistent facial features and unsteady motion, while the adapter achieves fluid transitions and a coherent depiction of the roaring action. In the “child biking through water” scenario, videos without the adapter suffer from artifacts and temporal inconsistencies, including distorted water reflections and unnatural motion. The adapter addresses these issues effectively, creating smooth biking dynamics and realistic water effects. Similarly, in the “walking dog” example, frame flickering and uneven body movements occur without the adapter, but are eliminated when it is used, producing smoother, more natural strides. Finally, in the “man surfing” illustration, the adapter strengthens semantic alignment by maintaining the surfer’s posture and interaction with the waves, resulting in visually cohesive and dynamic transitions. These case studies demonstrate how the adapter can improve temporal stability and semantic alignment, ensuring high-quality video outputs.

In Figures~\ref{fig:attention_maps} and~\ref{fig:attention_maps_bear} (all in supplementary material), we compare cross-attention maps generated by the base Stable Diffusion model and our adapter-enhanced model during the image-generation process. Minor differences in these attention maps underscore the adapter’s influence on the model’s performance.



\begin{figure}[H]
\centering
\includegraphics[width=0.92\linewidth]{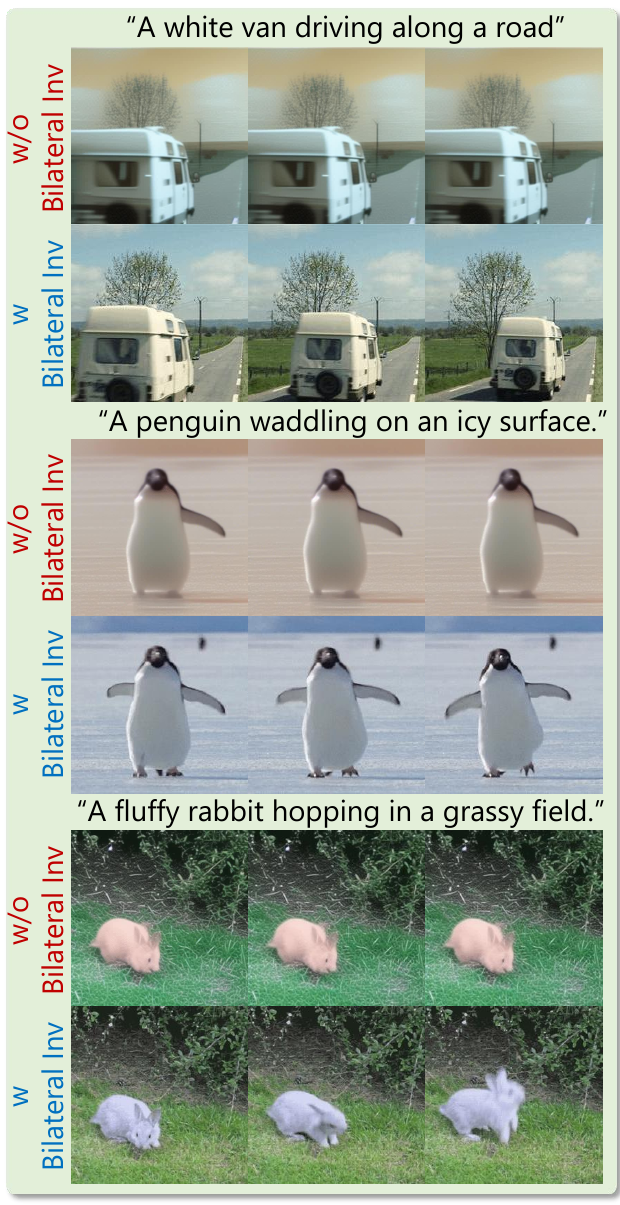}
\caption{Comparison of video generation results with bilateral filtering using different kernel sizes in the Stable Diffusion pipline.}
\label{fig:bilateral_filtering}
\Description{}\end{figure}

\begin{figure*}[htbp]
    \centering
    \includegraphics[width=\textwidth]{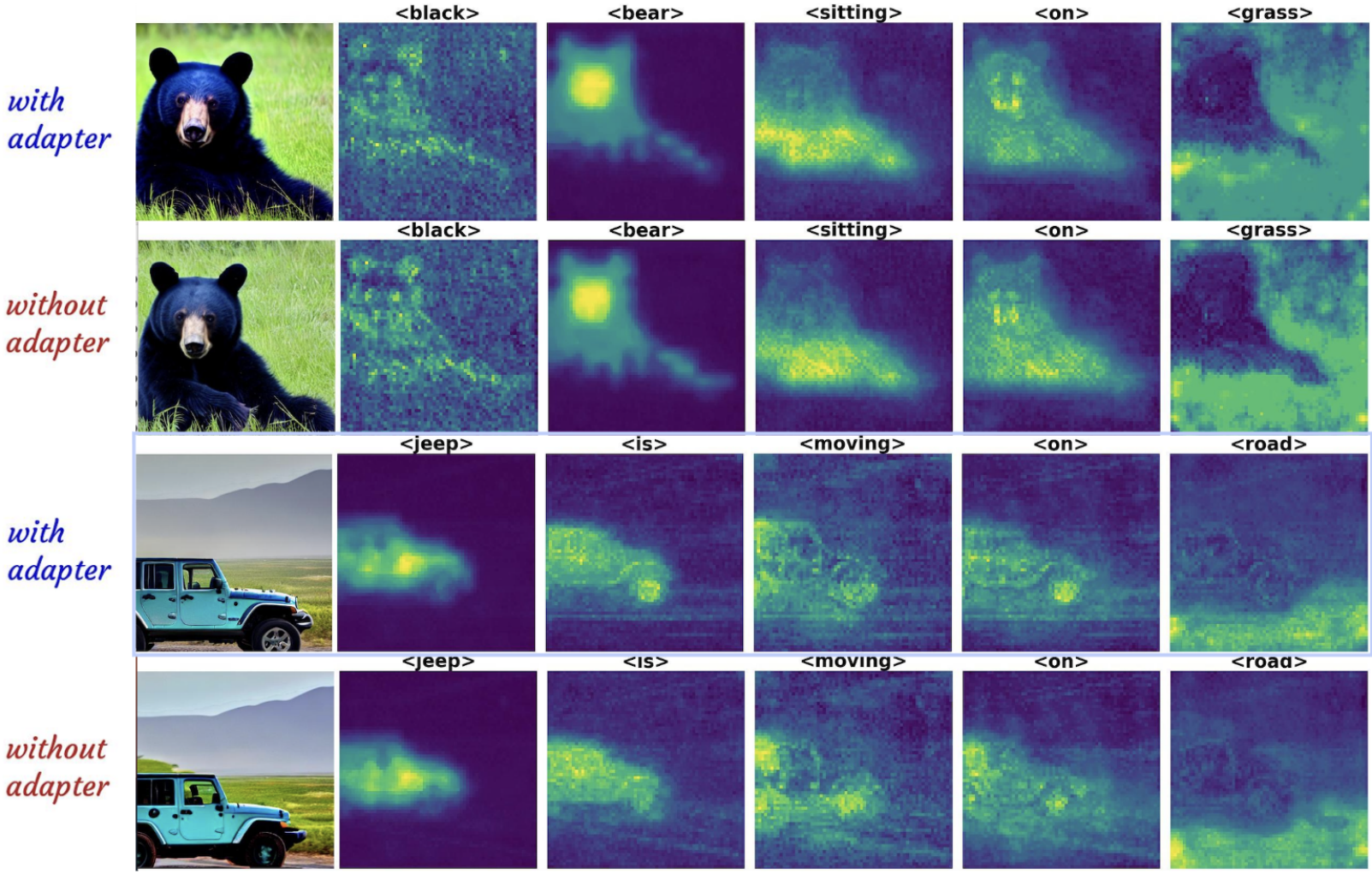}
    \caption{Visualization of attention maps comparing video frames generated with and without the adapter in the Stable Diffusion 1.5 pipeline.}
    \label{fig:attention_maps_bear}
\Description{}\end{figure*}

\begin{figure*}[htbp]
    \centering
    \includegraphics[width=0.96\textwidth]{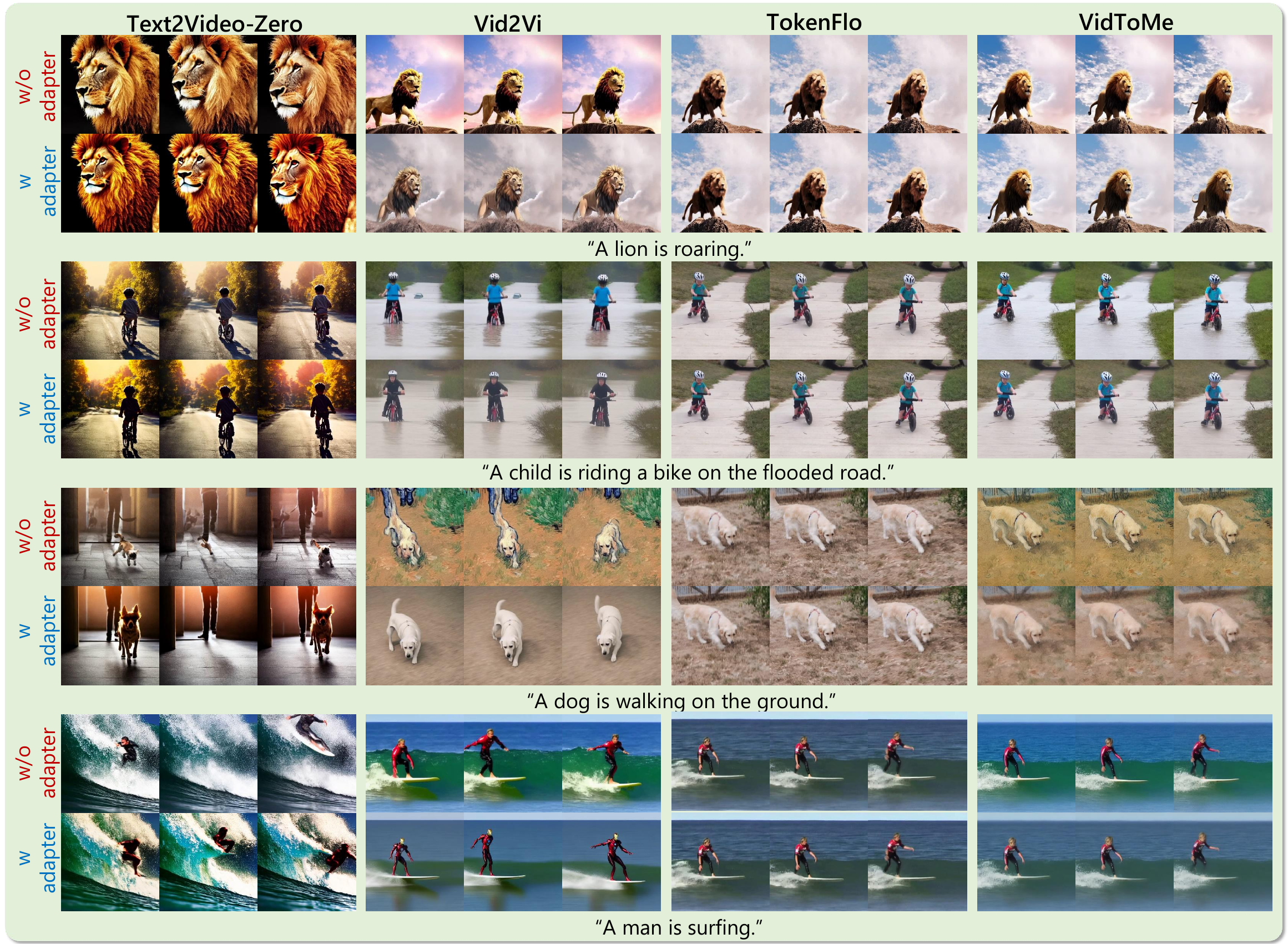}
    \caption{A comparison of video generation quality across different algorithms, with and without the use of an adapter.}
    \label{fig:lora_comparison}
    \Description{}
\end{figure*}

\begin{figure*}[htbp]
    \centering
    \includegraphics[width=\textwidth]{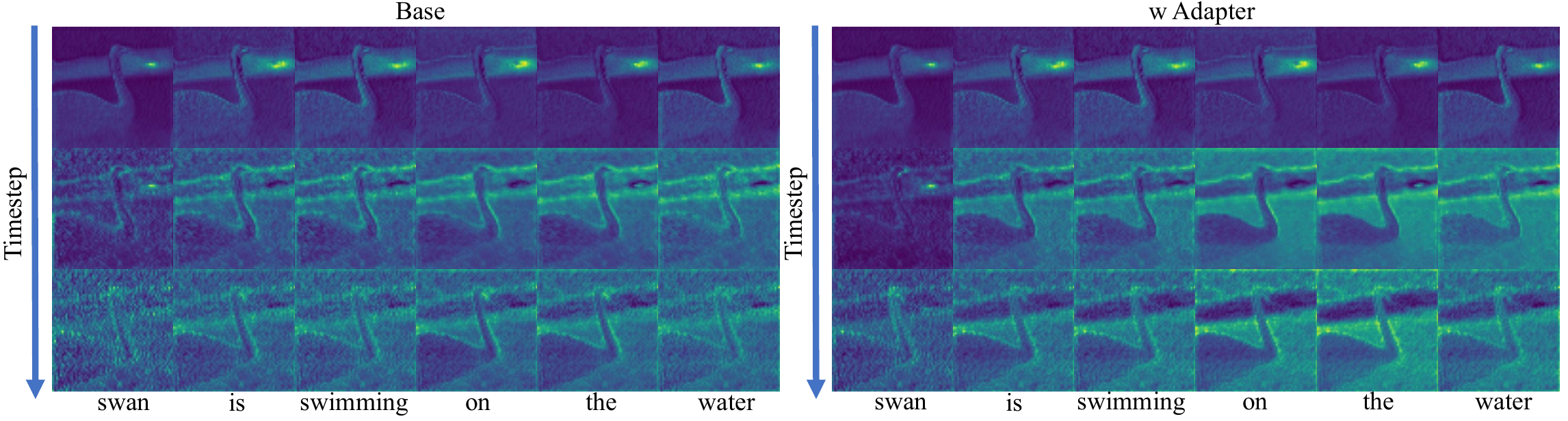}
    \caption{Visualization of attention maps from Unet's third upsampling block, comparing base and adapter models across 1000 timesteps. The corresponding timesteps from top to bottom are 1, 541, and 981.}
    \label{fig:attention_maps}
\Description{}\end{figure*}


\begin{figure*}[htbp]
    \centering
    \includegraphics[width=\linewidth]{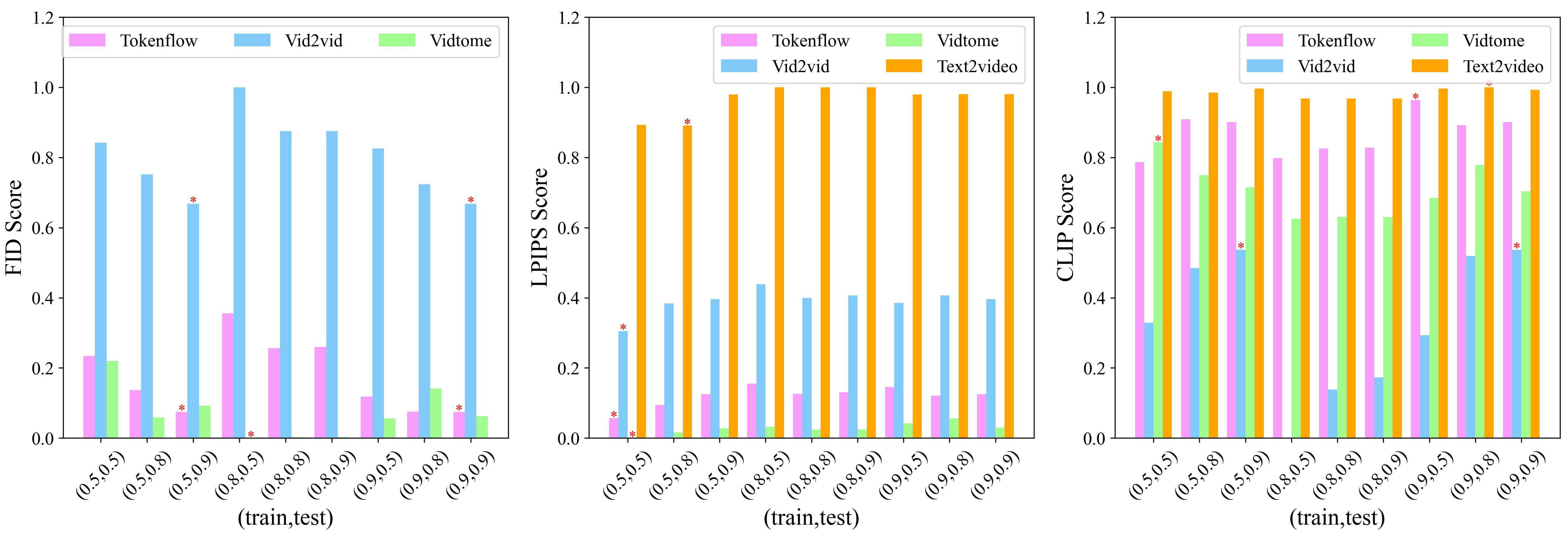}
    \caption{Bar chart illustrating the performance of four algorithms across three metrics for various combinations of (training timesteps, inference timesteps). The values 0.5, 0.8, and 0.9 represent the ranges 0.5t–1.0t, 0.8t–1.0t, and 0.9t–1.0t, respectively, where t denotes 1000 timesteps. The number on the left of the tuple indicates the training timesteps, while the number on the right represents the inference timesteps. Note that Text2Video, as a text-to-video generation algorithm, lacks pre-edited videos and therefore does not have an FID metric. The best (training timesteps, inference timesteps) combination for each algorithm is marked with an asterisk (*) at the top of the corresponding bar.}\label{fig:ts}
\Description{}
\end{figure*}

\begin{figure*}[htbp]
    \centering
    \includegraphics[width=\linewidth]{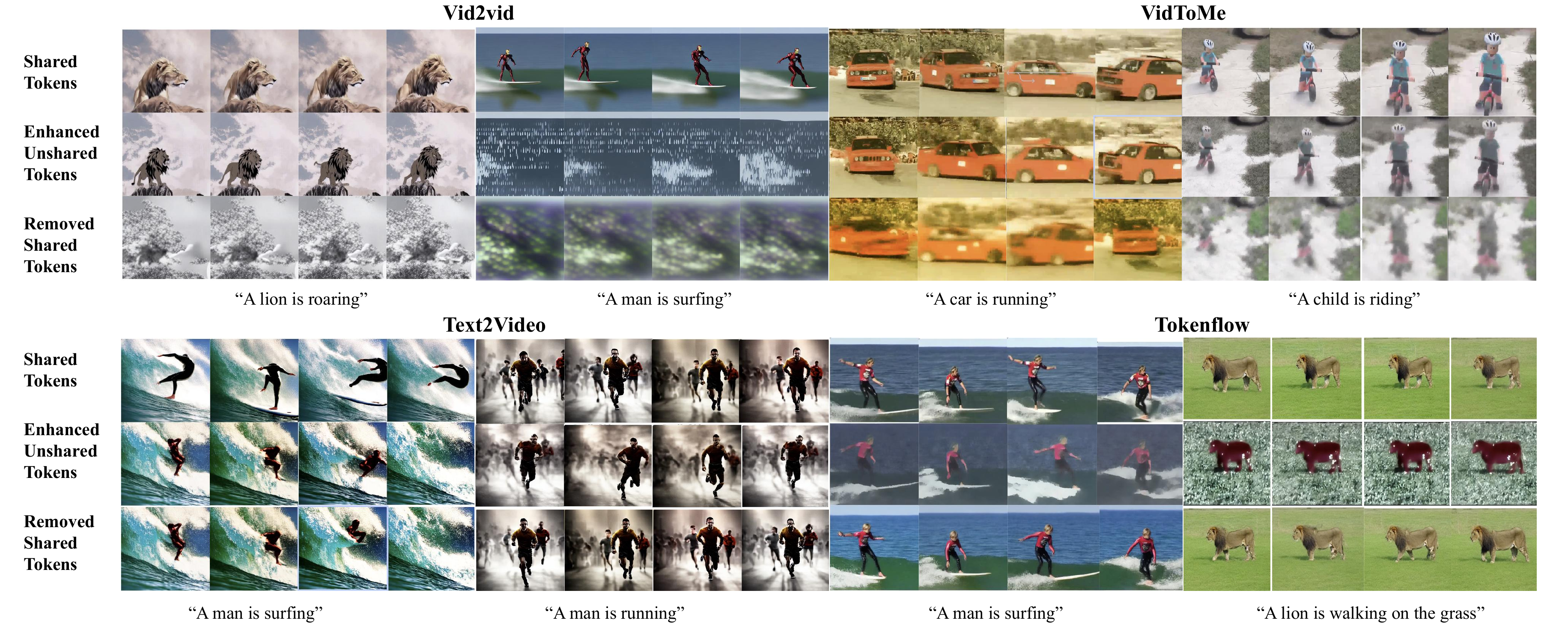}
    \caption{Visualization of video generation models (Vid2Vid, VidToMe, Text2Video, and TokenFlow) using different token strategies: Shared Tokens, Enhanced Unshared Tokens, and Removed Shared Tokens, evaluated across multiple prompts.}
    \label{fig:token_ablation}
\Description{}\end{figure*}

\begin{figure*}[htbp]
    \centering
    \includegraphics[width=0.95\textwidth]{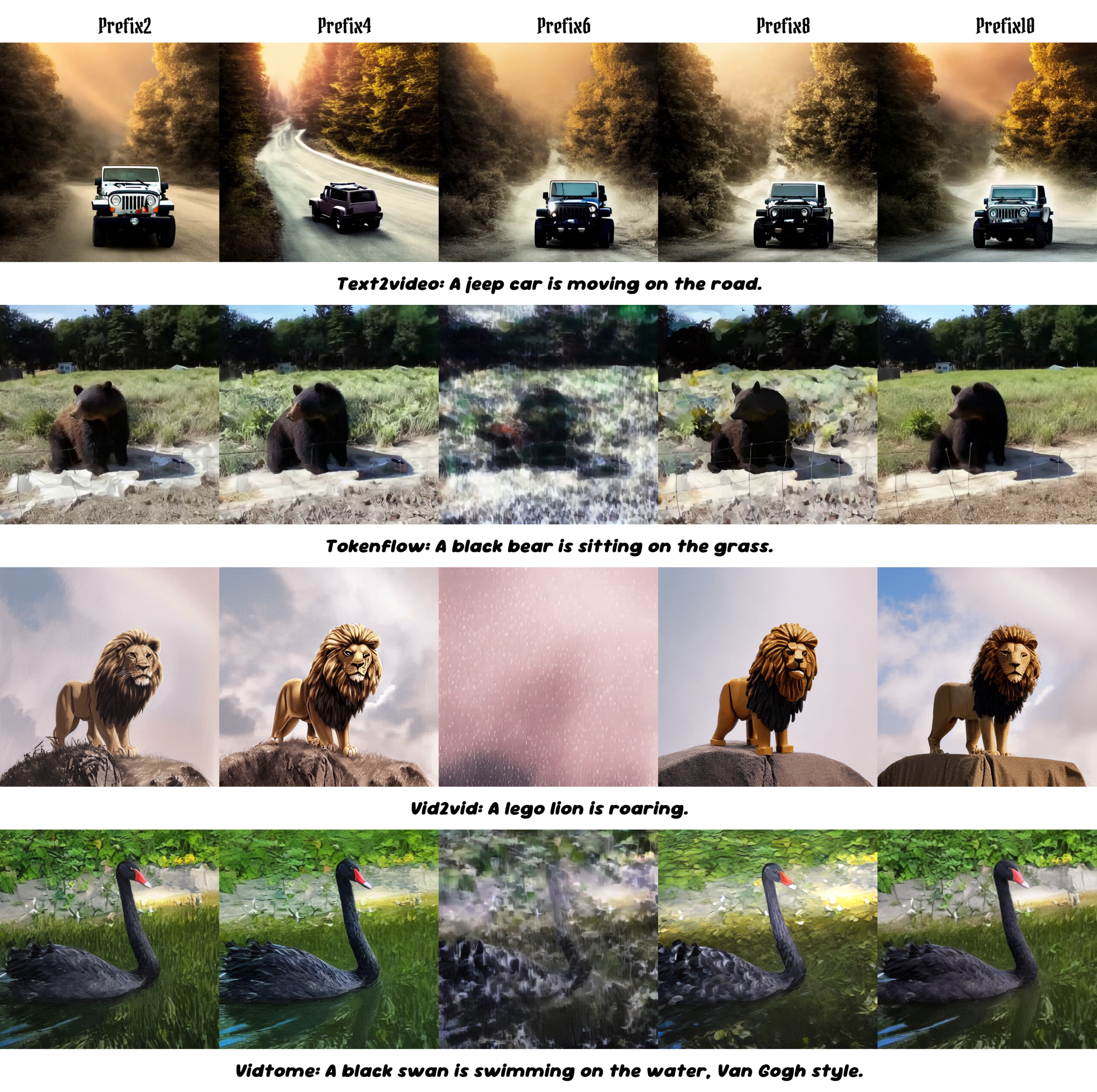}

    \caption{Visualization results obtained by Prompt learner taking different number of shared token (2, 4, 6, 8, 10) training and inference.}
    \label{fig:token_appendix}
\Description{}\end{figure*}

\begin{figure*}[htbp]
    \centering
    \includegraphics[width=\linewidth]{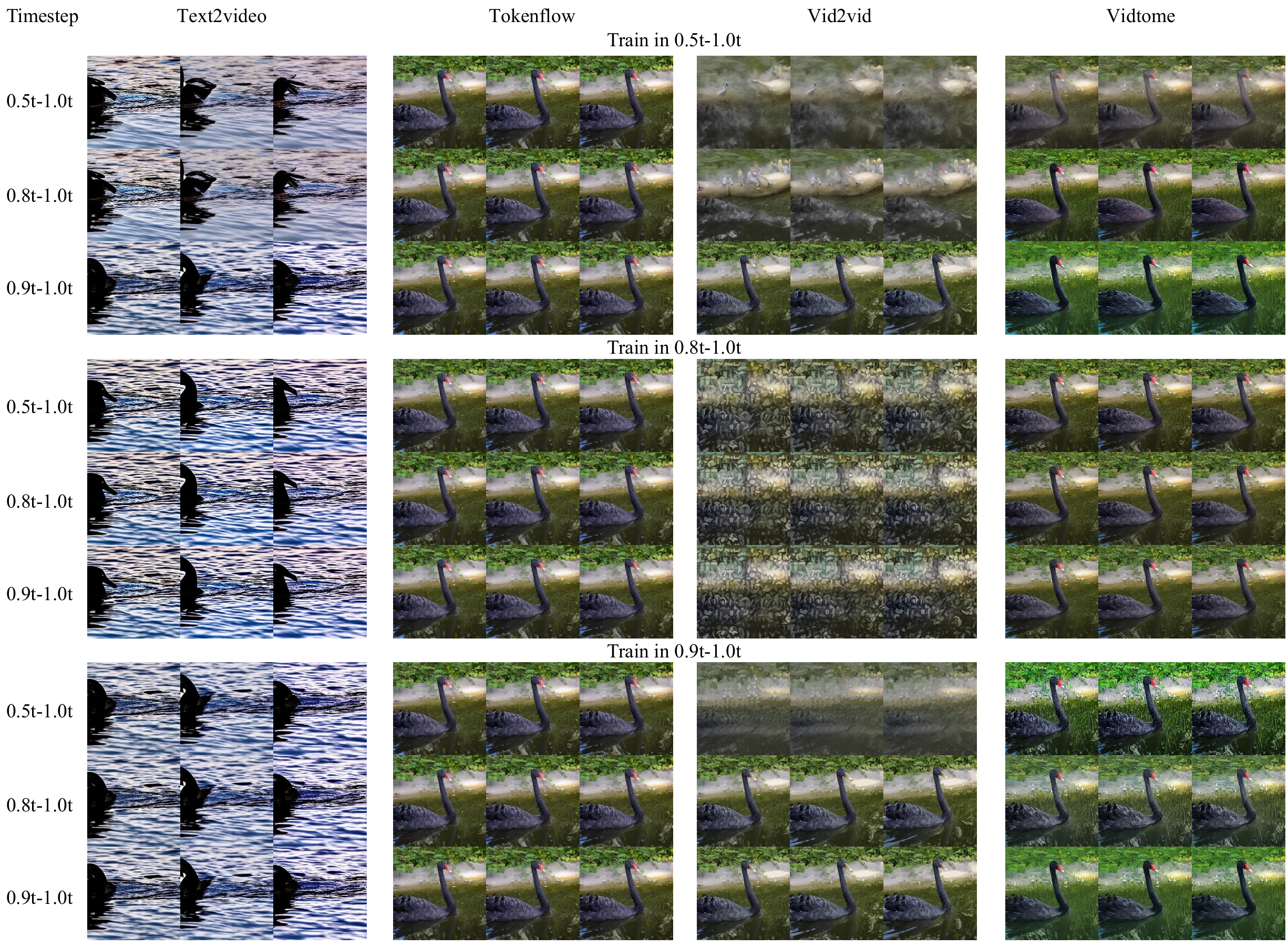}
        \caption{Visual comparison of a 0.5-1.0t training range for temporal-aware loss with a 0.9-1.0t adapter activation in inference range versus setting both to 0.5t (t = 1000).}
    \label{fig:stepvis}
\Description{}\end{figure*}

\begin{figure*}[htbp]
    \centering
    \includegraphics[width=\textwidth]{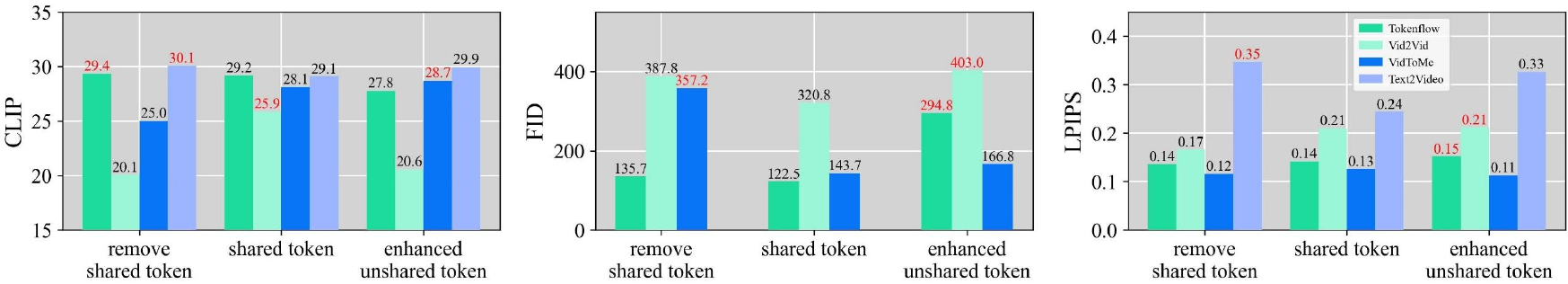}
    \caption{Comparison of token configurations—removing shared tokens, using shared tokens, and using enhanced unshared tokens—based on LPIPS, CLIP, and FID metrics.}
    \label{fig:token_comparison}
\Description{}\end{figure*}

\end{document}